\documentclass[10pt]{article} 
\usepackage[preprint]{rlc}

\usepackage{amssymb}            
\usepackage{mathtools}          
\usepackage{mathrsfs}           
\usepackage{graphicx}           
\usepackage{subcaption}         
\usepackage[space]{grffile}     
\usepackage{url}                

\usepackage{algorithm}
\usepackage{algpseudocode}
\usepackage{float}

\usepackage{booktabs}
\usepackage{multirow}

\usepackage{todonotes}

\title{External Model Motivated Agents: Reinforcement Learning for Enhanced Environment Sampling}

\author{Rishav Bhagat  \\
    rbhagat8@gatech.edu \\
    Georgia Institute of Technology
    \And
    Jonathan Balloch \\
    balloch@gatech.edu\\
    Georgia Institute of Technology
    \And
    Zhiyu Lin\\
    zhiyulin@gatech.edu\\
    Georgia Institute of Technology
    \And
    Julia Kim\\
    julia.kim@gatech.edu\\
    Georgia Tech Research Institute
    \And
    Mark Riedl\\
    riedl@cc.gatech.edu\\
    Georgia Institute of Technology
}

\begin{document}

\maketitle

\begin{abstract}
Unlike reinforcement learning (RL) agents, humans remain capable multitaskers in changing environments.
In spite of only experiencing the world through their own observations and interactions, people know how to balance focusing on tasks with learning about how changes may affect their understanding of the world. 
This is possible by choosing to solve tasks in ways that are interesting and generally informative beyond just the current task. 
Motivated by this, we propose an agent influence framework for RL agents to improve the adaptation efficiency of external models in changing environments without any changes to the agent's rewards.
Our formulation is composed of two self-contained modules: interest fields and behavior shaping via interest fields. 
We implement an uncertainty-based interest field algorithm as well as a skill-sampling-based behavior-shaping algorithm to use in testing this framework. 
Our results show that our method outperforms the baselines in terms of external model adaptation on metrics that measure both efficiency and performance.
\end{abstract}

\section{Introduction}
\label{sec:introduction}

Deep reinforcement learning (RL)~\citep{sutton2018introduction} has proven useful in solving tasks in many fields, including playing games~\citep{mnih2015human,silver2016mastering}, robotics~\citep{gu2017robotmanipulation}, and traditional control applications such as power management~\citep{mao2016resource,zhang2019power}. 
In some of these applications, the RL agent whose primary objective is defined with task rewards may additionally need to explicitly model phenomena not directly related to that primary task. 
These models that train on data acquired while the agent performs the task, which we refer to as ``external models'', are often present in complex systems such as robots or smart devices.

External models are used to estimate descriptive statistics, distinguish safe vs unsafe areas, and model environment dynamics, and can be used to inform third-parties like humans or prepare the agent itself for downstream tasks. 
Consider an autonomous underground mining robot whose primary task is to dig and search for minerals, but that is also being used to model and monitor whether the environment is safe for humans, looking for dangers such as toxic air or intense temperatures~\citep{nanda2010application}. 
Although a temperature safety model may not be relevant to the robot's mining task since temperatures can pose a danger to humans long before the robot, it is critical to human safety that the temperature model remains accurate.  
While many use-cases might not require the task policy and an external model to be trained simultaneously, if an environment transfer occurs during deployment, such as moving to a new depth underground or damage to the robot, both the task policy and external model must be quickly adapt to the change.

These external models generally exist outside of the reinforcement learning loop and therefore have no conventional way to dictate to the policy what new data is needed. In the case of environment change, we hypothesize that explicitly identifying and acquiring more informative data for external model adaptation is crucial to improving the performance and efficiency of the external model.
While 
non-greedy behavior in RL is conventionally used to compensate for the shortcomings of greedy policy learning, we argue that, in the presence of some environment change,  intelligent agents supporting external models will need to use non-greedy behavior to  update those models as well. 
Inaccurate external models could result in increased risk if high-stakes decisions, such as exploring an area the model predicts is safe, are made using them.
Thus, this work examines how an RL agent in a changing environment can balance updating its task policy with improving the efficiency and performance of external model adaptation. 

The key idea for this work is that there are \textit{points of interest} that, if observed, would benefit adaptation of the external model.
Our main contributions are the following: (1) a framework to motivate agents in a task reward agnostic way using two modules: a definition of ``interest'' and a method to steer agents by leveraging interest; (2) an exemplar implementation of both modules to motivate the agent with the secondary objective of enhanced environment sampling for external model learning; (3) experiments and results demonstrating that our interest-based agent motivation implementations improve the adaptation of external models in the presence of environment change compared to the standard PPO and online DIAYN baselines.

\section{Related Work}

\paragraph{Multi-Objective Reinforcement Learning.}

The core technical challenge of this work focuses on influencing the behavior of an RL agent in a changed environment according to a model with an objective unrelated to the agent's reinforcement task. 
As single-objective RL is poorly suited to handle multiple objectives scenarios like these~\citep{hayes2022practical}, multi-objective reinforcement learning (MORL) is often used to optimize one or more policies according to multiple objectives simultaneously~\citep{roijers2013survey,van2014multi}. 
Our work is related to, but different from, multi-objective RL because our secondary objective of improving external model adaptation is not an RL task.
As a result, existing multi-objective RL methods do not fit this problem.  

Influencing agent behavior according to a complex set of contraints or objectives, however, is not limited to MORL. Policy shaping~\cite{NIPS2013_e034fb6b} enables an oracle policy to influence a learned policy, while reward shaping~\citep{ng1999policy,knox2009interactively} and intrinsic rewards~\citep{chentanez2004intrinsically,aubret2019survey,linke2020adapting} steer agent behavior by changing the reward function with factors separate from the task. However, we do not assume access to an oracle policy, and reward shaping only has a delayed impact on the policy and often produces brittle policies that are highly sensitive to changes in the environment~\citep{fu2018learning}. 
%
%
Goal-conditioned~\citep{nair2018visual,liu2022goalrlsurvey,hafner2022deep} and skill-conditioned~\citep{eysenbach2018diversity,sharma2019dynamics} reinforcement learning has become a popular alternative that influences behavior by learning a policy conditioned on a certain goal or skill vector. 
Related to options in hierarchical RL~\citep{sutton1999between,stolle2002learning,pateria2021hierarchical}, by learning an embedding space of skills or goals, policy conditioning can be used to immediately influence agent behavior depending on how the goal or skill is selected. 
As such, our work uses skill-conditioned policies to trade-off objectives for our unique multi-objective problem setting.  

\paragraph{Interesting Points for Learning.}
In this work, skills are used to direct the policy toward samples that are most beneficial, i.e. ``interesting,''  to the external model's transfer learning process. 
Adapting to the changed environment with transfer learning allows the model to improve learning efficiency by reusing knowledge from the source task~\citep{transferlearning}. 
Different domains have different definitions of what makes a sample \textit{interesting}. 
For example, in the supervised learning setting \textit{active learning} examines how to increase learning sample efficiency by labeling only the most ``interesting'' unlabeled dataset samples~\citep{ren21active}. 
However, in online, agent-based learning like reinforcement learning, there is no dataset of unlabeled samples; an agent can only train its models using samples it observes from interacting with the environment. 
To make reinforcement learning more sample efficient, exploration methods use uncertainty estimates, similar to the methods used in active learning, as well as metrics such as sample complexity, model error, and state-space coverage to identify and direct agents to more interesting parts of the environment~\citep{tarbouriech21samples,pathak19disagreement, an2021edac, yao21sample, ren21active}. 
Inspired by this variety of interest definition, we define our behavior influence function to be independent of interest definition, and in our exemplar interest implementation we take inspiration from the body of work that uses uncertainty.

\section{Approach}\label{sec:approach}

In this work, we developed an approach, which we call \textbf{External Model Motivated Agent} (EMMA), to influence an agent towards more interesting samples for an external model while performing its task. 
The EMMA approach formulates the solution to this unique multi-objective problem as a modular framework for reward-agnostic agent motivation, described in Section~\ref{sec:approach:framework}. 
Additionally, we designed exemplar implementations of the framework's interest and influence modules, which are described in Sections \ref{sec:mc_droput} and \ref{sec:poi_diayn}.

\begin{figure}
    \centering
    \includegraphics[width=\textwidth]{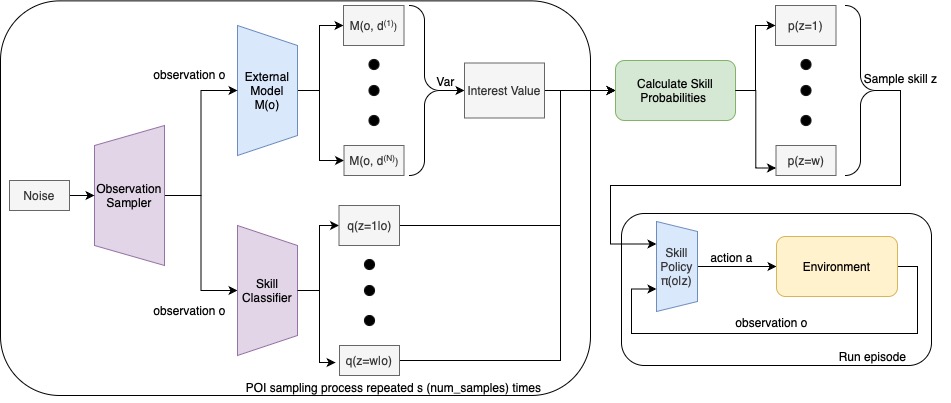}
    \caption{This system diagram outlines the process that occurs per episode to shape the behavior of a skill-based agent via the Monte Carlo dropout disagreement interest field. 
    The POI sampling process occurs at the beginning of each episode $s$ times and then once a skill is sampled it is fixed for the full episode. 
    After a rollout of data (with many episodes) are generated, the external model, policy, and other models introduced for our methods are updated using the new data.}
    \label{fig:method_diagram}
\end{figure}

\subsection{Reward-Agnostic Agent Motivation Framework}\label{sec:approach:framework}

Our agent motivation framework consists of two separate modules: Point of Interest Field (Interest Field) and Point of Interest Influence (POI Influence). The \textit{interest field} is a scalar field over the observation space \(O\) that defines how ``interesting'' an observation is for the policy to visit. The \textit{POI influence} module uses the interest field to shape the agent's behavior to collect more ``interest'' throughout the on-policy rollout. Together, these modules form a task reward-agnostic motivation framework. 
Figure~\ref{fig:method_diagram} shows the complete framework and illustrates how the Interest Field and POI Influence modules interact.





The first module is the interest field: $ f_{POI}: O \rightarrow \mathbb{R} $. This field defines how interesting a given observation is to the external model.
For our methods, we require the scalar field to be defined at every observation in the observation space, allowing it to be queried for any observation $o \in O$, irrespective of the agent's current state. These constraints form a standard interface for the POI influence module to use that allows for global task-agnostic interest information to be queried. 




Once an interest field is calculated across the observation space, it is still not trivial to shape agent behavior using such a field for a few reasons. The interest field usually has nothing to do with the task and may provide conflicting goals for the agent to perform. The interest field may often be highly non-stationary, so using models to estimate a ``value'' for the interest field in an RL sense may be challenging, as they would be chasing a moving target. Balancing these goals of providing information to the external model while still learning to complete the task is an essential part of our method. The goal of any of our POI influence algorithms is to maximize the interest our agent ``collects'' as it learns to complete the task at hand. 

Recalling the earlier example of an autonomous underground mining robot modeling safety for humans, our approach would use reward-agnostic behavior shaping to guide the agent towards areas that provide informative samples to the safety model after environment changes. 
Upon reaching a deeper level, where the temperature distribution changes and temperature concerns become more prevalent, a framework like the EMMA would enhance the human safety model by being able to define areas of unexplored temperature readings as interesting and then influence the robot to execute its task near those high-interest locations.

\subsection{Exemplar Interest Field: Monte Carlo Dropout Disagreement}
\label{sec:mc_droput}

The primary goal of our agent motivation strategy is to enhance the information gain of the external model $M$ throughout policy episodes. 
Fundamentally, we can maximize the information gained by the external model by minimizing the uncertainty in the predictions~\citep{mackay1992info}. 
As such, areas of high model uncertainty are areas that the agent should prioritize visiting.

Various methods exist for quantifying model uncertainty \citep{ABDAR2021243}. For our experiments, we employ Monte Carlo dropout \citep{dropout}, a technique suited for parameterized external models. This method involves estimating uncertainty by computing the variance of model predictions using multiple iterations of dropout. Formally, the interest field $f_{POI}: O \rightarrow \mathbb{R}$ for an observation $o$ is defined as:
$$ f_{POI}(o) = \mathrm{Var}[M(o, d^{(1)}), M(o, d^{(2)}), \ldots, M(o, d^{(N)})] $$
where $d^{(i)}$ represents the $i$-th dropout sample and $N$ denotes the number of dropout samples used. This function can be computed for any observation $o$ without the agent currently observing $o$ or having access to any ground truth information about the underlying state, as required. If the model has multiple outputs, then the variance across each output should be computed and then the variances should be summed or averaged.

It is worth noting that while Monte Carlo dropout serves as our uncertainty quantification method, alternative uncertainty metrics \citep{ABDAR2021243} can be seamlessly incorporated into our methodology without compromising its effectiveness.

\subsection{Exemplar POI Influence: Interest-Valued Discrete Skill Sampling}
\label{sec:poi_diayn}

Consider using a discrete skill-conditioned agent $\pi_{\theta}(o | z)$ along with a skill classifier $q_{\phi}(z|o)$ that classifies any state to one of a finite set of skills. 
This agent can be directed towards paths with higher interest by biasing the skill sampling every episode. 
Using a VAE \citep{Kingma2014} to sample observations (see Appendix \ref{obs_sampling}) these observations can then be passed to the skill classifier. 
The interest value is computed for each of these observations and assigned to the skill that was classified for that observation. 
Then a skill is sampled from the set based on the average interest per observation of the skills. 
We introduce $\eta$ to weight a uniform prior along with the interest biased prior to improve task convergence across the skills. This algorithm for biased skill sampling per episode is outlined in Algorithm \ref{alg:poi_skills}.

\begin{algorithm}[h]
\caption{POI Influence using an Interest Biased Discrete Skill Prior}
\label{alg:poi_skills}
\begin{algorithmic}[1]
\Require uniform prior weight $\eta$, num skills $w$, num samples $s$ 
    \State Sample $s$ observations $X \in O^s$ from the observation sampler
    \State Calculate POI: $I_x = f_{POI}(x) \in \mathbb{R} \quad \forall \quad x \in X$
    \State Estimate skills posteriors: $Q_{(z|x)} = q_{\phi}(z | x) \quad \forall \quad z \in [1, \dots, w], x \in X $
    \State For each skill, calculate the average interest per step: $A_z = \cfrac{\sum_{x \in X} Q_{(z|x)} \cdot I_x}{\sum_{x \in X} Q_{(z|x)}} \quad \forall \quad z \in [1, \dots, w]$
    \State Calculate probabilities: $P_z = \text{Softmax}(A_z) \quad \forall \quad z \in [1, \dots, w]$
    \State Sample skill $z \sim p(z)$ where $p(z) = \frac{\eta}{w} + (1 - \eta)P_z$
    \State Proceed for episode using skill $z$
\end{algorithmic}
\end{algorithm}

The DIAYN (Diversity Is All You Need) algorithm \citep{eysenbach2018diversity} learns a skill-conditioned policy in which it incentivizes diversity by rewarding the agent when a discriminator predicts the current skill well using the current observation. This encourages the skills the policy learns to be more diverse and cover more of the observation space. We use DIAYN in an online fashion, in which the DIAYN reward is added to the extrinsic reward. This should incentivize the agent to find solutions that are seperable in observation space by their skill. In our experiments, we use online DIAYN (with the discriminator as our skill classifier) as our skill-conditioned policy learning process since it generates diverse skills, but there is no reason our techniques couldn't be applied to any other skill-based policy paired with a skill classifier. It should be noted that in DIAYN intrinsic reward, we assume a uniform prior as this part reward is only used as reward normalization. We use POI DIAYN to refer to our skill sampling technique paired with online DIAYN.

\section{Experiments}
\label{sec:exp_setup}

\paragraph{Procedure.}

In our experiments, we train an RL agent using a chosen interest field and POI influence algorithm along with proximal policy optimization (PPO) as the backbone RL algorithm. 
After some predefined number of steps (unknown to the agent), an environment transfer is injected. The agent and the external model must both adapt to the new post-transfer environment as efficiently as possible.
The external model $M$ and the policy $\pi$ are both trained on the rollout data at the end of a rollout. For the external model, we train 8 epochs per rollout; we found that increasing the number of epochs per rollout for the model greatly increased the sample efficiency of the method (see Appendix~\ref{app:exps} for a sensitivity study of epochs per rollout). 


\paragraph{Environment.}

We use a task  from NovGrid \citep{balloch2022novgrid} called \textbf{DoorKeyChange}, a grid world environment in which there exists one red key, one blue key, and a red locked door with the goal past the door. Pre-transfer, the red key opens the red door as expected; post-transfer, the blue key opens the red door, and the red key stops working. 

\paragraph{External Model Specification.}
\label{sec:ext_models}


For our experiments, we define the external model specification as \textit{correct key distance prediction}, in which a neural network is trained to predict the Euclidean distance to the key that currently opens the door. If the key is not within the agents view, the model should predict $14$ (a distance that ensures the key is outside the agent's view). This model is a good model to test our algorithms since the transfer specified impacts the outputs of this model and efficient adaptation of this model could potentially be used to aid agent adaptation.

\paragraph{Metrics.}
\label{sec:def_metrics}

We use two metrics to measure the effectiveness of our methods in external model adaptation to environment transfer. \textit{Adaptive efficiency} is defined as the number of steps after the transfer the smoothed external model loss takes to reach a predefined convergence threshold. \textit{Adaptive performance} is the minimum smoothed external model loss after the transfer. The parameters for loss smoothing and convergence thresholds are listed in Appendix \ref{app:hypers}. The metrics are measured in on policy rollouts and random agent rollouts to provide a way to measure the external model on data most applicable to the task and external model generalization outside of on policy data. 

As RL is noisy, we use 10 different runs each with different random seeds for each of our experiments. 
We filter out runs that did not learn to complete the task since these runs would inflate the performance of the methods that do not converge on the task ($<20\%$ of runs for our methods and baselines do not converge). We use interquartile mean (IQM) to aggregate the metrics from converged runs since IQM produces more reproducible results~\citep{agarwal2021deep}.

\section{Results}
\label{sec:results}

We ran experiments with the setup discussed in the Section \ref{sec:exp_setup} (using hyperparameters shown in Appendix \ref{app:hypers}) using the \textit{correct key distance prediction} external model specification, to test if our interest based methods are more sample efficient and performant for external model learning post-transfer while also solving the task. These methods aim to have lower adaptive efficiency values (more efficient) and lower adaptive performance value (more performant) compared to the PPO and online DIAYN baselines. For ease of comparison, we normalize these metrics by PPO performance. 

We present results for the Monte Carlo dropout uncertainty interest fields along with the following POI influence algorithms: POI DIAYN and POI intrinsic reward with global POI embedding (included for comparison, see Appendix \ref{sec:ir_emb} for details). We compare these methods with two POI-free baselines: Vanilla PPO and Online DIAYN with PPO. This experiment aims to show how these different POI influence algorithms perform with uncertainty-based interest compared to our baselines. Our POI DIAYN method converged to the task pretty consistently (more than 80\% of the time), so we were able to use an $\eta$ of 0.

\begin{figure}[h!]
    \centering
    \subcaptionbox{On policy rollout external model losses.\label{fig:8_train_em_losses}}[0.5\textwidth]{
        \includegraphics[width=\linewidth]{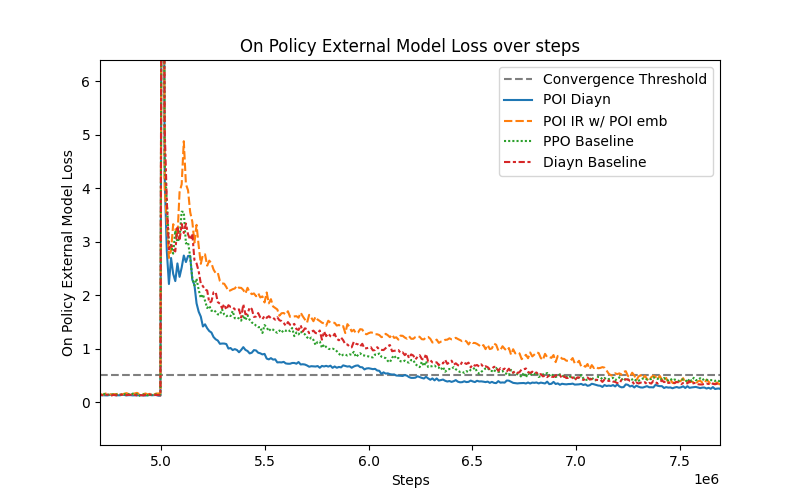}
    }\hfill
    \subcaptionbox{Random agent rollout external model losses.\label{fig:8_eval_em_losses}}[0.5\textwidth]{
        \includegraphics[width=\linewidth]{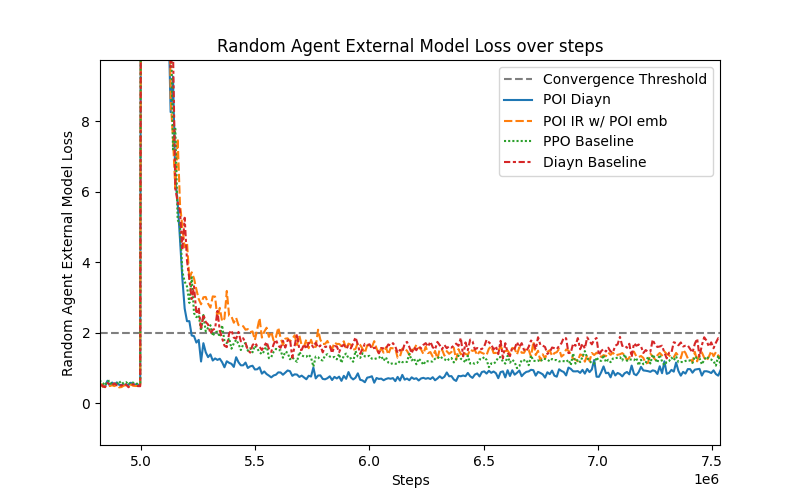}
    }
    \caption{These plots show the \textit{correct key distance} external model losses over environment steps on both on policy and random agent rollouts. The steps directly after the transfer are shown to highlight the impact of our algorithms in adapting the extenal model. 
    The plots show the IQM of external model loss for each method at each time step. In these plots, lower values are better as they correspond with better external model performance. These graphs will not align perfectly with the results in Figure \ref{fig:8_metrics} as these graphs aggregate losses per step while the table aggregates calculated metrics.}
    \label{fig:8_em_losses}
\end{figure}

\begin{figure}[h!]
    \centering
    \begin{tabular}{|l|c|c|c|c|}
        \hline
        \multirow{2}{*}{\textbf{Algorithm}} & \multicolumn{2}{c|}{\textbf{Adaptive Efficiency} $\downarrow$} & \multicolumn{2}{c|}{\textbf{Adaptive Performance} $\downarrow$} \\
        \cline{2-5}
        & On Policy & Random Agent & On Policy & Random Agent \\
        \hline
        POI DIAYN & \textbf{0.451220} & \textbf{0.945701} & \textbf{0.794670} & \textbf{0.684142} \\
        POI IR w/ POI emb & 1.019512 & 1.107692 & 0.980950 & 1.102312 \\
        DIAYN Baseline & 0.674390 & 1.006335 & 1.164005 & 1.239283 \\
        PPO Baseline & 1.000000 & 1.000000 & 1.000000 & 1.000000 \\
        \hline
    \end{tabular}
    \caption{This table shows the post-transfer metrics defined in Section \ref{sec:def_metrics} for the interest-based methods and non-interest baselines on the experimental setup described. The IQM of the converged runs is used to aggregate the 10 seeds we ran. The adaptive efficiency values are the normalized (by PPO performance) number of steps till the external model loss hits the convergence threshold. The adaptive performance values are the normalized (by PPO performance) minimum external model loss post transfer. Lower values are better for both these metrics.}
    \label{fig:8_metrics}
\end{figure}


The quantitative comparison of metrics in Figure \ref{fig:8_metrics} shows that the external model adapts most efficiently when using POI DIAYN, which can also be seen in the graphs in Figure \ref{fig:8_em_losses}. 
These results confirm our hypothesis that uncertainty-based interest paired with an interest biased skill sampling method yields better environment sampling for external model learning over both a pure task agent and a diversity-driven task agent.

\section{Discussion and Conclusions}

The experimental results described in Section \ref{sec:results} provide much insight into our interest-based methods for learning external models via agent rollouts. 
As noted in Figure \ref{fig:8_metrics}, interest-based methods yield performance gains in both sample efficiency and asymptotic performance post-transfer. 
The improvement in adaptive performance is especially apparent on the evaluations using the random agent rollouts; evaluations on the random agent are challenging for the baselines because the external model must generalize beyond just the on-policy greedy rollouts of the trained, RL task-only agent. 
By learning from preferred environment samples acquired by the interest-influenced agent, the external model likely receives the data it needs to generalize more efficiently over more of the observation space. 
This shows that the EMMA framework can be used to help external models influence agents for faster convergence. 
Furthermore, these results confirm that an interest field based on external model uncertainty estimates using Monte Carlo dropout disagreement is a good choice to motivate the agent, and that POI DIAYN can be used as a way to steer agent behavior towards high-interest areas.


In our work, we investigated whether using \textit{interest fields} to motivate an agent with the purpose of learning an external model more effectively yields better external model adaptation when faced with environment transfer in terms of sample efficiency and asymptotic performance. While our experiments focused on key distance prediction, the EMMA framework can be extended to a variety of external models including but not limited to world models, policy models, and safety prediction models (see Appendix \ref{app:external_models}). These models are widely used in reinforcement learning, particularly in environments where novelties can arise.  Our results demonstrate the potential of interest-driven behavior shaping to improve the adaptation of external models and open up avenues for iteration on our methods through the behavior shaping framework.

Our findings suggest that interest-driven methods can serve as a powerful tool for shaping agent behavior agnostic to rewards. In our use-case, we motivated the agent to aid in the learning of an external model, but our algorithms work with any definition of interest. This flexibility opens avenues for a wide range of investigations into the interactions between interest signals and agent behavior. For example, using recency-based interest may yield more patrolling behaviors to visit stale states. We aim for this work inspires future research to further explore the applications of this paradigm, leading to more versatile and adaptable agents in complex environments with transfers.

\bibliography{main}
\bibliographystyle{rlc}

\appendix

\section{Learning Repeats Sensitivity Study}\label{app:exps}

This sensitivity study examines the effects of varying the number of epochs per rollout on the relative efficiency of different interest-driven methods and baselines. 

\subsection{4 epochs per rollout}
\label{sec:main_exp}

In addition to our results with 8 epochs per rollout, we also ran an experiment using the same setup described in Section \ref{sec:exp_setup} with only 4 epochs per rollout. As seen by Figure \ref{fig:metrics}, the interest based methods outperform the baselines. However, interestingly our secondary method described in Appendix \ref{sec:ir_emb} does better than POI DIAYN in this experiment. Nevertheless, POI DIAYN still outperforms the baselines in all of our metrics by a wide margin even with 4 epochs per rollout.


\begin{figure}[ht!]
    \centering
    \begin{tabular}{|l|c|c|c|c|}
        \hline
        \multirow{2}{*}{\textbf{Algorithm}} & \multicolumn{2}{c|}{\textbf{Adaptive Efficiency} $\downarrow$} & \multicolumn{2}{c|}{\textbf{Adaptive Performance} $\downarrow$} \\
        \cline{2-5}
        & On Policy & Random Agent & On Policy & Random Agent \\
        \hline
        POI DIAYN & 0.783172 & 0.915441 & 0.779286 & \textbf{0.604408} \\
        POI IR w/ POI emb & \textbf{0.627832} & \textbf{0.862745} & \textbf{0.747689} & 0.694266 \\
        DIAYN Baseline & 0.994606 & 1.029412 & 0.938544 & 1.044733 \\
        PPO Baseline & 1.000000 & 1.000000 & 1.000000 & 1.000000 \\
        \hline
    \end{tabular}
    \caption{This table shows the same metrics as in Figure \ref{fig:8_metrics}, except now for the experiment with only 4 epochs per rollout.}
    \label{fig:metrics}
\end{figure}

\subsection{1 epoch per rollout}

We briefly present the results of an experiment with the same setup as in Section \ref{sec:exp_setup}  except now only using 1 epoch per rollout of data. With only one epoch of training per rollout in the environment, it is likely that the model did not fully learn everything it could from that rollout of data. Thus, our methods will likely not work as well.

\begin{figure}[h!]
    \centering
    \begin{tabular}{|l|c|c|c|c|}
        \hline
        \multirow{2}{*}{\textbf{Algorithm}} & \multicolumn{2}{c|}{\textbf{Adaptive Efficiency} $\downarrow$} & \multicolumn{2}{c|}{\textbf{Adaptive Performance} $\downarrow$} \\
        \cline{2-5}
        & On Policy & Random Agent & On Policy & Random Agent \\
        \hline
        POI DIAYN & 1.137800 & 0.965402 & 1.006910 & \textbf{0.829252} \\
        POI IR w/ POI emb & 1.028761 & \textbf{0.951451} & 1.036848 & 0.912862 \\
        DIAYN Baseline & 1.079646 & 1.008929 & \textbf{0.950950} & 0.904382 \\
        PPO Baseline & \textbf{1.000000} & 1.000000 & 1.000000 & 1.000000 \\
        \hline
    \end{tabular}
    \caption{This table shows the same metrics as in Figure \ref{fig:8_metrics}, except now for the experiment with only a single epoch per rollout.}
    \label{fig:1_metrics}
\end{figure}

As seen in Figure \ref{fig:1_metrics}, the performance gains of our methods are lost when only using a single epoch per rollout. This result aligns with the expectation as each rollout is not nearly fully learned, so actively seeking specific rollouts will not help.

\subsection{Analysis}

The discrepency of the results using different values for epochs per rollout is unsurprising as the enhanced environment sampling provided by our algorithms can be better used when more of the information from each sample is captured. If a sample is not fully learned before it is thrown out, then the next sample doesn't necessarily need to contain new information in it to optimally aid in external model training as even giving the same sample again would result information gain to the external model.

\section{Impact on Task Performance}
\label{sec:task_performance}

While the main motivation for our work is to improve the external model performance while the agent is learning a task, it is important to note on the changes to task performance when applying external model motivated agents.

\begin{figure}[ht!]
    \centering  
    \subcaptionbox{8 epochs per rollout.\label{fig:8_rewards}}[0.32\textwidth]{
        \includegraphics[width=\linewidth]{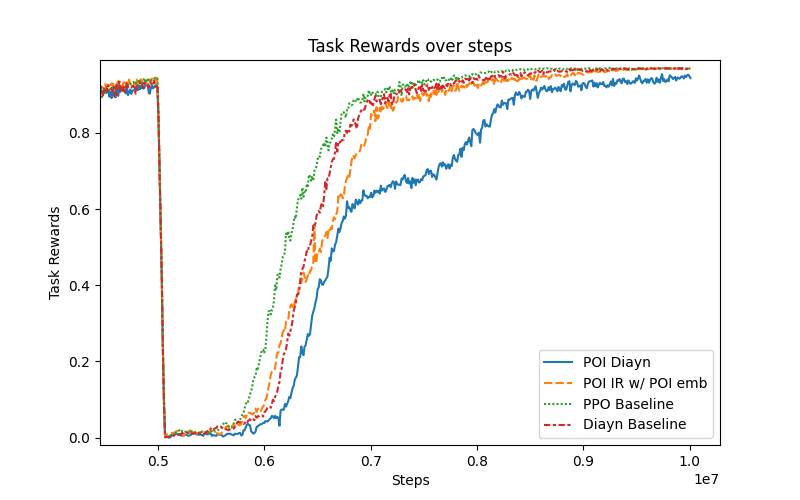}
    }
    \subcaptionbox{4 epochs per rollout.\label{fig:4_rewards}}[0.32\textwidth]{
        \includegraphics[width=\linewidth]{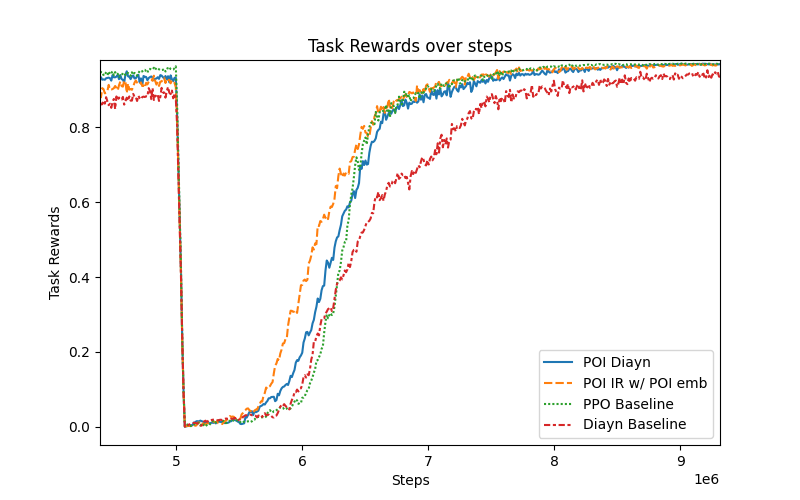}
    }
    \subcaptionbox{1 epoch per rollout.\label{fig:1_rewards}}[0.32\textwidth]{
        \includegraphics[width=\linewidth]{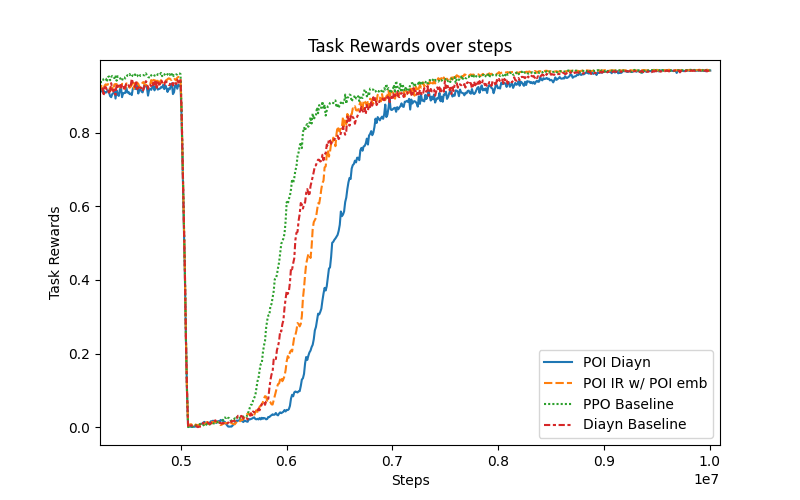}
    }
    \caption{This plot shows the task performance of all the methods with the transfer occuring at 5M steps. As shown in the plot, task performance drops significantly at the transfer and then quickly rises again. This plot uses interquartile mean values of the 10 runs for each algorithm at each time step. For this plot, higher values indicate better performance on the task (Door Key Change).}
    \label{fig:reward_graph} 
\end{figure}

Figure \ref{fig:reward_graph} shows the reward learning curves for each of the methods in each experiment post-transfer. As seen by the graphs, our methods do not impact task learning too much as in all of the experiments included within the interquartile mean, the agent converged on the task. Our experiments ran with around at least an 80\% convergence rate with POI DIAYN and at least a 70\% with POI instrinsic reward w/ a global POI embedding. Additionally, the graphs show that the efficiency of the agent is impacted to some extent by using a diverse skill based method like DIAYN, but this is more due to the diverse nature of online DIAYN rather than our contributions. 

These results are also quite noisy because if the interquartile seeds end up including one poorly performing run (as we see for POI DIAYN in Figure \ref{fig:8_rewards} and DIAYN in Figure \ref{fig:4_rewards}), the IQM significantly gets decreased. Some loss of performance is expected as the agent is now attempting to optimizing two potentially competing objectives of solving the task while also learning the external model well, and these results show that the loss of performance on the task is not too much.

\section{Additional External Model Specifications}
\label{app:external_models}

We provide code to run experiments using the correct key distance specification or any of the external models listed in this section (see Appendix \ref{app:code}) on a variety of tasks. Further, these additional external model specifications listed below are not unique to MiniGrid (unlike the one used in our main experiments), or even discrete action spaces, so they can be run with Mujoco or any other environment.

\paragraph{Policy Model.}

Setting the external model in our algorithms to the policy model effectively converts all of our methods into traditional exploration methods. To calculate uncertainty of the policy model, we must add dropout to the policy and use the different dropout samples as an ensemble the same way our other external models were used. We used the same dropout hyperparameters used for correct key distance prediction. Additionally, to calculate the disagreement of a skil conditioned policy, we need a skill in addition to the observation. To solve this problem, we use the skill classifier and condition the policy with the skill that is most probable if the agent was in the given observation.

\begin{figure}[h!]
    \centering
    \includegraphics[width=0.7\linewidth]{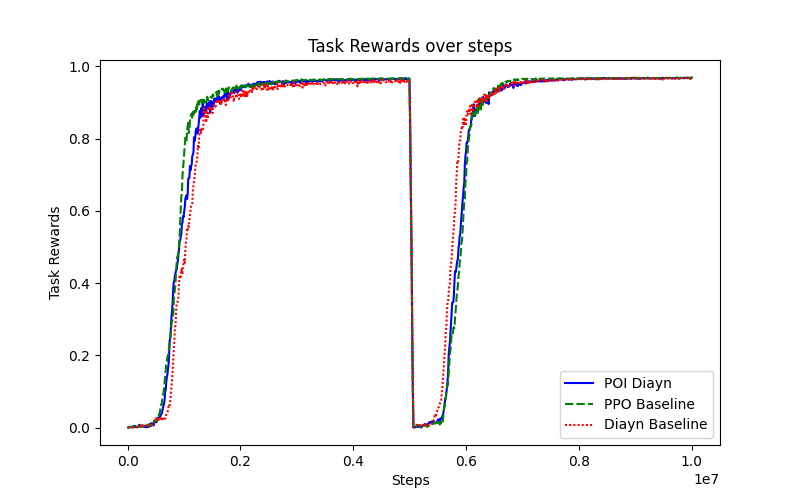}
    \caption{These plots are showing the IQM of the task reward (on door key change) when using our methods with the policy model set as the external model. For this plot, higher values imply better task performance and are thus prefered. Just like previous experiments, the transfer is injected at step 5M, after which the agent must adapt to the new environment. All three methods depicted here perform similarly.}
    \label{fig:policy_poi_rewards}
\end{figure}

We present some preliminary results in \ref{fig:policy_poi_rewards} using this interest-based exploration method on the door key change task with higher learning repeats (16) than used on the experiments shown previously in this work (4) (see Appendix \ref{app:hypers} for a full list of hyperparameters).

With this specification, task performance becomes the single objective and our interest methods are pure exploration methods, thus we evaluate our methods using task rewards. Our POI DIAYN method does not show significant improvement over the two baselines provided in this work. It should also be noted that improvement over vanilla PPO is less of an accomplishment in the context of exploration since there already exists plenty of methods that outperform vanilla PPO in terms of exploration \citep{RE3, RISE, GIRM, REVD, ladosz2022exploration}. While this interest field didn't show improvement with POI DIAYN, another potential interest field could be the entropy of the policies output given an observation since this is another indicator for uncertainty. This removes the need for using a dropout ensemble for the policy. Additionally, further tuning for using policy models as external models likely needs to be done to use our methods in this way.

\paragraph{Forward Dynamics Model.}
Another possible external model specification is a forward dynamics model that takes the current observation and an action and then predicts the next observation. It should be noted that our algorithms presume the model takes only an observation, but this can be remedied by producing an action that makes the most sense to use when calculating interest. For our skill based method, we pass the observation through the policy to produce the action as that is what the agent will do at that observation. For the intrinsic reward based methods, we use the action that the agent actually took for the reward.

Learning a forward dynamics model efficiently is something prior work in model-based reinforcement learning has attempted \citep{pathak19disagreement, yao21sample}; however, this is still an active field of research. Further, these methods generally do not specifically aim for solely learning good dynamics models as a separate objective and instead aim for optimal policy performance, differentiating them from our methods.

\paragraph{Imputation Models}

Another external model that would be useful in agent learning and adaptation, is an imputation model that fills in any missing observation values. Consider a scenario in which a robot that uses noisy sensors for observation is attempting to perform a task. An imputation model to fill in these missing values would immensely aid  in the learning of the task. Whenever an observation is received, self-supervised masking can be used to generate ground truth values to the imputation model to train on. Specifically, this model would take a partial observation as input and output a full observation that could be passed directly to the policy. Motivating an agent to adapt this model quickly would likely also aid in adapting the agent since any missing values being correctly filled in would make the task easier for the agent.

\section{Additional Methods}
\label{sec:additional_methods}

Here we present some additional methods that overall performed worse that POI DIAYN, but may still be more applicable for other use cases. Further, these methods likely can be developed further to perform well on tasks that are more tightly specified.

\subsection{Interest Fields}

\subsubsection{Model Parameter Gradient Fields}

Another proxy that can be used for information gain is the amount the model changes from seeing an observation. Consider basic gradient descent on a single observation $o$ paired with its ground truth $y$ for an external model $M_{\theta}: O \rightarrow \mathbb{R}^d$ parameterized by $\theta \in \mathbb{R}^p$. The general update rule looks like (with learning rate $\alpha$),

$$ \theta \leftarrow \theta - \alpha \nabla_{\theta} L(y, M_{\theta}(o) $$

So, the model parameter change is $\alpha||\nabla_{\theta} L(y, M_{\theta}(o)||$. For the following derivation, we use $\nabla_{\vec{x}}$ to denote the vector gradient (of shape $ \text{length of } \vec{x} \times 1$) and $J_{\vec{x}}$ to denote the Jacobian matrix of vector valued function (of shape $\text{length of } f \times \text{length of } \vec{x} $). By chain rule,

$$ \nabla_{\theta} L(y, M_{\theta}(o)) = J_{\theta}[M_{\theta}(o)]^T \nabla_{M_{\theta}(o)} L(y, M_{\theta}(o)) $$

Now, by the Cauchy-Schwartz inequality \citep{wu2009various},

$$ ||\nabla_{\theta} L(y, M_{\theta}(o))|| \leq ||J_{\theta}[M_{\theta}(o)]^T||_F ||\nabla_{M_{\theta}(o)} L(y, M_{\theta}(o))|| = ||J_{\theta}[M_{\theta}(o)]||_F ||\nabla_{M_{\theta}(o)} L(y, M_{\theta}(o))|| $$

where $||A||_F$ denotes the Frobenius norm of $A$ and $||\vec{x}||$ is the euclidean norm of $\vec{x}$. Thus, the upper bound of the model change is proportional to $||J_{\theta}[M_{\theta}(o)]||_F$, which only depends on $\theta$ and $o$, so we can set this to the interest function without breaking the constraints of being able to calculate it while the agent does not have access to $y$.

$$f_{POI}(o) = ||J_{\theta}[M_{\theta}(o)]||_F$$

While potential model change seems like a decent indicator for interest, it can be quite slow to calculate as it requires a gradient (backward) calculate over the network for each calculation of $POI$. Thus, it should not be used along with a POI influence algorithm that makes frequent calls to the $POI$ function. For this reason, the focus of our experiments are on other interest fields.

\subsubsection{Transfer Aware Fields}

For certain tasks, there may be some priors or assumptions that can be made about the novelties that may occur in the environment. Consider knowing that each step there is a probability $p$ that a transfer occurs in a specific position in the environment. Then, it may be useful to use a "staleness" interest field that incentivises the agent to feed the external model the recent data for the full position space. 

Due to the generic framework and separation of the interest field and the POI influence algorithm that uses this field, task specific interest fields can be designed and easily integrated with our exploration methods.

\subsection{Behavior Shaping via Interest Fields}

\subsubsection{Intrinsic Rewards}

One simple method for POI influence via these points of interest is to use the interest value as an intrinsic reward signal. There is a plethora of prior work on designing intrinsic rewards that optimize exploration for policy learning \citep{burda2018exploration, RISE, GIRM, REVD, RE3}; however, we directly use interest values as these values motivate the agent to optimize for external model learning. That is, we reformulate the reward signal to be:

$$ r = r^{(i)} + r^{(e)} $$

where $r^{(e)}$ is the task reward provided by the environment and $r^{(i)} = f_{POI}(o)$ where $o$ is the observation that the agent is observing while receiving this reward.

This method is relatively straightforward in terms of both implementation and theory, but it has a few problems. One potential drawback of the fact that this intrinsic reward may be highly non-stationary making it non-trivial \citep{pmlr-v119-cheung20a} for the RL algorithm to learn to optimize for this $POI$ without any sense of global $POI$ information across the observation space. Another drawback of this method is that it requires a call to the $POI$ function every step, which can be computationally costly.

\subsubsection{Intrinsic Rewards w/ an interest embedding}
\label{sec:ir_emb}

One of the main drawbacks to just using the $POI$ signal as an intrinsic reward is that the agent will likely learn too slow to adapt to changes in the $POI$ signal, which can be extremely dynamic depending on the definition of interest. Thus, it may be useful for the policy $\pi: O \rightarrow A$, where $A$ is the action space of the agent, to be conditioned on some sort of global $POI$ information. However, for continuous or large observations spaces, it is not possible to just enumerate the possible observations and their respective interest values. Thus, we attempt to construct a vector $\vec{e}_{POI}$ that contains some sort of semi-global interest information while maintaining the same intrinsic reward from above.

\paragraph{Observation Space Sampling.} \label{obs_sampling}

First, as the observation space may be large, we need a way to sample observations that can be directly passed into the interest field to produce $POI$ values. In our experiments, we used a variational autoencoder (VAE) \citep{Kingma2014}, with encoder $E_o: O \rightarrow \mathbb{R}^\ell$ and decoder $D_o: \mathbb{R}^\ell \rightarrow O$ where $\ell$ is the size of the latent space. After each rollout of the policy, this VAE is trained using the variation ELBO loss. Although we use VAEs, any observation sampler can work, so if one is provided with the environment it should be used. Further, sampling states from a replay buffer is also a valid option.

\paragraph{Embedding Learning.}

Now, we attempt to create an embedding $\vec{e}_{POI}$ that contains the information about the global state of interest across the observation space. Logically, if a model can "query" this vector with an observation and produce the $POI$ of that observation, then the information is contained in that vector (or only depends on the obserservation and wont change over time). 

Thus, we construct two deep models: an embedding update model $U: O^s \times \mathbb{R}^s \times \mathbb{R}^q \rightarrow \mathbb{R}^q$ (where $q$ is the size of the $POI$ embedding and $s$ is the size of the set of observations and pois) and a $POI$ prediction model $P: O \times \mathbb{R}^q \rightarrow \mathbb{R}$. The embedding update model is trying to embed the information of the observations and their $POI$ into the current $POI$ embedding. Meanwhile, the prediction model is effectively attempting to decipher the information within the embedding to extract the $POI$s, showing that the embedding genuinely contains the $POI$ information. Additionally a few eval only observation + $POI$ pairs that were not given to the update model are applied to the loss calculated at the end of the prediction model to encourage the models to generalize beyond only the samples seen by the update model.

Thus, to train our models,

\begin{algorithm}[h]
\caption{POI Embedding Update and Prediction Model Training}
\begin{algorithmic}[1]
\Require Current POI embedding $\vec{e}_{POI}$, num samples $s$, num eval samples $s_e$
\State Sample $s$ observations $X \in O^s$ from the observation sampler
\State Sample $s_e$ observations $X_e \in O^{s_e}$ from the observation sampler
\State Calculate POI: $POI = f_{POI}(X)$
\State Calculate POI for eval observations: $POI_e = f_{POI}(X_e)$
\State Update POI embedding: $\vec{e}_{POI}' = U(X, POI, \vec{e}_{POI})$
\State Predict POI: $\hat{POI} = P(X, \vec{e}_{POI}')$
\State Predict POI for eval observations: $\hat{POI}_e = P(X_e, \vec{e}_{POI}')$
\State Calculate loss: $L = \text{MSE}(\hat{POI}, POI) + \text{MSE}(\hat{POI}_e, POI_e)$
\State Update models $U$ and $P$ to minimize $L$
\end{algorithmic}
\end{algorithm}

Now, to generate/update our current $POI$ embedding

\begin{algorithm}[h]
\caption{Update $\vec{e}_{POI}$}
\begin{algorithmic}[1]
\Require Current POI embedding $\vec{e}_{POI}$, num samples $s$
\For{$i = 1$ to $k$}
    \State Sample $s$ observations $X \in O^s$
    \State Calculate POI: $POI = f_{POI}(X)$
    \State Update POI embedding: $\vec{e}_{POI} = U(X, POI, \vec{e}_{POI})$
\EndFor
\end{algorithmic}
\end{algorithm}

\paragraph{Policy Conditioning.}

Finally, the policy $\pi$ can now be conditioned using $\vec{e}_{POI}$. The conditioned policy $\pi: O \times \mathbb{R}^q \rightarrow A$ has access to global interest information across the full observation space. Consider, an agent that has to make a decision of whether to follow the left path or right path in a maze (assume both lead to the goal in equal number of steps). The agent should follow the left path if there is more interest on the left while it should follow the right path otherwise. Since the interest field can easily change a lot after a rollout, this information should be encoded into $\vec{e}_{POI}$ and then passed to agent to help make the decision.




\section{Experiment Hyperparameters}
\label{app:hypers}

In this appendix we provide the set of hyperparameters used to generate our results. Any parameters not provided here were ommited for brevity, but can be found in the configs folder within the code implementation (see Appendix \ref{app:code}). We held any parameters that were agnostic to our methods constant between the different methods to allow for fair comparisons.

\begin{figure}[h!]
    \centering
    \begin{tabular}{|l|l|}
        \hline
        \multicolumn{2}{|c|}{\textbf{DoorKeyChange}} \\
        \hline
        pre\_transfer\_steps & 5000000  \\
        post\_transfer\_steps & 5000000 \\ 
        \hline
        \multicolumn{2}{|c|}{\textbf{PPO}} \\
        \hline
        learning\_rate & 0.00075 \\
        n\_steps & 2048 \\
        batch\_size & 256 \\
        n\_epochs & 4 \\
        gamma & 0.99 \\
        gae\_lambda & 0.95 \\
        clip\_range & 0.2 \\
        ent\_coef & 0.05 \\
        vf\_coef & 0.5 \\
        max\_grad\_norm & 0.5 \\
        policy\_layers\_sizes & [980, 256, 64, 7] \\
        value\_layers\_sizes & [980, 256, 64, 1] \\
        \hline
        \multicolumn{2}{|c|}{\textbf{Correct Key Distance Prediction}} \\
        \hline
        layer\_sizes & [980, 100, 10, 1] \\
        layer\_activations & ReLU \\
        dropout\_p & 0.5 \\
        num\_mc\_dropout\_samples & 30 \\
        batch\_size & 256 \\
        \hline
        \multicolumn{2}{|c|}{\textbf{DIAYN}} \\
        \hline
        num\_skills & 5 \\
        beta & 5 \\
        discriminator\_layer\_sizes & [980, 200, 5] \\
        final\_discriminator\_activation & Softmax \\
        other\_activations & LeakyReLU \\
        discriminator\_batch\_size & 128 \\
        \hline
        \multicolumn{2}{|c|}{\textbf{POI DIAYN}} \\
        \hline
        num\_samples\_for\_poi\_calc $s$ & 1000 \\
        \hline
        \multicolumn{2}{|c|}{\textbf{Observation Sampler}} \\
        \hline
        latent\_dim & 32 \\
        encoder\_layer\_sizes & [980, 200, 100, 64] \\
        decoder\_layer\_sizes & [32, 100, 100, 980] \\
        final\_decoder\_activation & Sigmoid \\
        other\_activations & LeakyReLU \\
        learning\_rate & 0.001 \\
        batch\_size & 256 \\
        \hline
        \multicolumn{2}{|c|}{\textbf{Metrics}} \\
        \hline
        loss\_smooting\_ewma\_span\_in\_rollouts & 30 \\
        on\_policy\_loss\_convergence\_threshold & 0.001 \\
        random\_agent\_loss\_convergence\_threshold & 0.5 \\
        \hline
    \end{tabular}
    \caption{This table provides the set of important hyperparameter values for reproducing our results. Additional hyperparameters can be found in the config files of the code implementation.}
\end{figure}

\section{Code Implementation}
\label{app:code}

We additionally provide the code used to generate the results in our work in \href{https://github.com/rishavb123/EMMA}{this repository}. Our implementation mainly uses the \href{http://pytorch.org}{PyTorch}, \href{https://stable-baselines3.readthedocs.io/en/master/}{Stable Baselines 3} \citep{raffin2021stable}, \href{http://gymnasium.farama.org}{Gymnasium}, and \href{https://pypi.org/project/experiment-lab/}{ExperimentLab} libraries to run our experiments.

\end{document}